\documentclass{article} 
\usepackage{iclr2019_conference,times}

\usepackage{amsmath,amsfonts,bm}

\def\eqref#1{equation~\ref{#1}}

\def\1{\bm{1}}

\DeclareMathAlphabet{\mathsfit}{\encodingdefault}{\sfdefault}{m}{sl}
\SetMathAlphabet{\mathsfit}{bold}{\encodingdefault}{\sfdefault}{bx}{n}

\usepackage{url}
\usepackage{graphicx}

\title{Using Videos to Evaluate Image Model Robustness}

\author{Keren Gu\thanks{Equal contribution. Work done as members of Google AI Residency program.}, Brandon Yang\footnotemark[1], Jiquan Ngiam, Quoc Le \& Jonathon Shlens  \\
Google, Mountain View, CA, USA\\
\texttt{\{kerengu, bcyang, jngiam, qvl, shlens\}@google.com} \\
}

\iclrfinalcopy 
\begin{document}

\maketitle

\begin{abstract}
Human visual systems are robust to a wide range of image transformations that are challenging for artificial networks. We present the first study of image model robustness to the minute transformations found across video frames, which we term "natural robustness". Compared to previous studies on adversarial examples and synthetic distortions, natural robustness captures a more diverse set of common image transformations that occur in the natural environment. Our study across a dozen model architectures shows that more accurate models are more robust to natural transformations, and that robustness to synthetic color distortions is a good proxy for natural robustness. In examining brittleness in videos, we find that majority of the brittleness found in videos lies outside the typical definition of adversarial examples (99.9\%). Finally, we investigate training techniques to reduce brittleness and find that no single technique systematically improves natural robustness across twelve tested architectures. 

\end{abstract}

\begin{figure}[h]
\begin{center}
\includegraphics[width=\columnwidth]{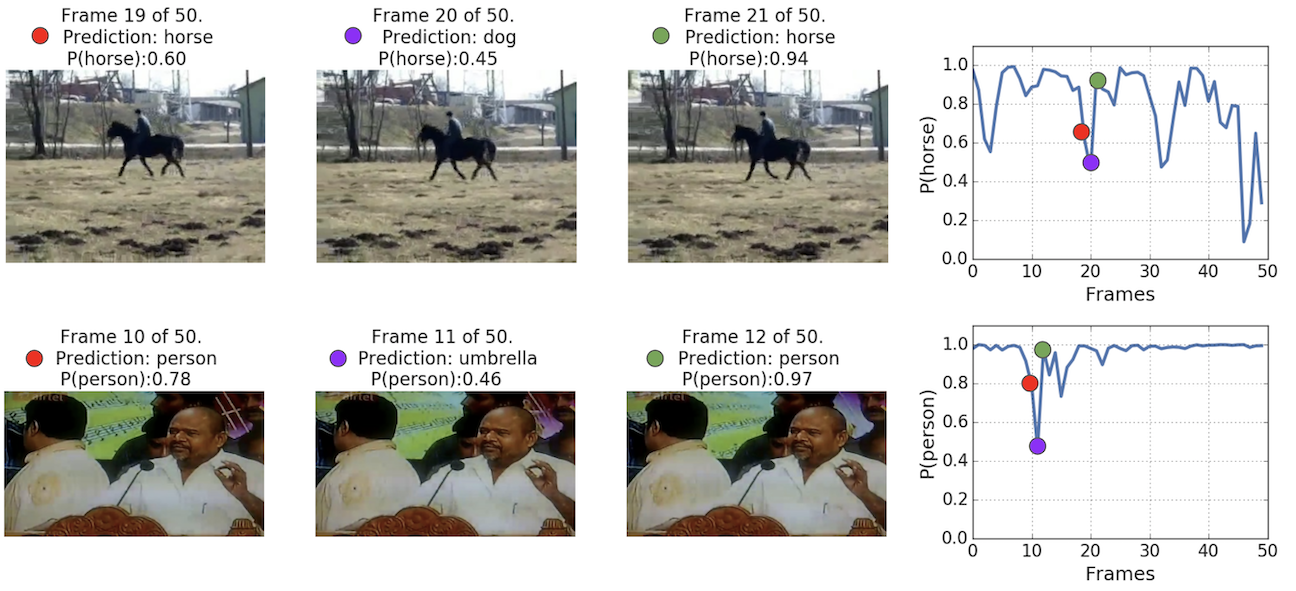}
\end{center}
\caption{
{\bf Deep convolutional image models are sometimes brittle to minute natural transformations across consecutive video frames.} We illustrate examples of a ResNet-50 (\cite{resnet-he2016deep}) model (85.5\% top-1 accuracy) producing varying predictions across consecutive video frames at 15Hz with visually imperceptible differences from the YouTube-BoundingBox (\cite{real2017youtube}) dataset. Note: these examples are hand selected to highlight instances of model brittleness.
}
\label{fig:context}
\end{figure}

\section{Introduction}
\label{intro}

Convolutional neural network (CNN) models have surpassed human performance on image classification benchmarks like Imagenet (\cite{he2015delving, deng2009imagenet}). However, their predictions are often sensitive to small image transformations (\cite{szegedy2013intriguing, polar_bear}) that are visually imperceptible to humans. In contrast, human visual systems are robust to a wide variety of distortions, including color and noise distortions (\cite{geirhos2018generalisation}). Our goal is to systematically study this phenomenon of model brittleness to small image transformations to better understand and improve the robustness of computer vision models.

Previously, model brittleness has commonly been studied through the framework of adversarial examples, where example images with minute but carefully constructed modifications can fool an otherwise accurate model (\cite{szegedy2013intriguing,kurakin2016adversarial,carlini2017adversarial}). 

Recent work has found that models are also brittle to examples that are not adversarially constructed. For instance, models are found to be sensitive to changes in camera settings such as exposure and contrast (\cite{temel2018traffic}), small translations and rotations (\cite{polar_bear, engstrom2017rotation}), and imperceptible variations across consecutive video frames (\cite{polar_bear}). Convolutional models are much more brittle than humans to synthetic distortions when the same distortions are not used during training. However, models trained on specific distortions can outperform humans on those specific distortions (\cite{geirhos2018generalisation}). To better evaluate the extent of model brittleness in the non-adversarial setting, the community has proposed datasets to benchmark robustness (\cite{hendrycks2018benchmarking} and \cite{temel2018traffic}). These proposed datasets mimic real world scenarios using increasingly more realistic synthetic distortions, such as perturbing clean images with artificial fog and snow. 

We present the first study of CNN robustness to natural transformations found across nearby video frames. We term this "natural robustness". This is an important type of robustness, because it models a diverse range of transformations that are closer to changes that occur in the natural environment than the synthetic distortions previously studied. We provide a framework, with dataset and evaluation metrics, to systematically evaluate natural robustness. Our results show that
\begin{enumerate}
\item More accurate image model architectures are more robust to natural evaluations. 
\item Small translations and synthetic color distortions are good proxies for evaluating natural robustness. 
\item No single regularization technique systematically improves natural robustness across model architectures.
\end{enumerate}

\section{Robustness Metrics}\label{sec:metric}

Given a correct classification on one frame, we define natural robustness as the conditional accuracy on the neighboring frame, similar to the Jaggedness measure (\cite{polar_bear}). More formally, let $f:X \rightarrow Y$ be a model mapping from input space $X$ to prediction space $Y$, and $d:X \rightarrow X$ defines an image transformation function. The robustness of a model $f$ is defined as $R_d(f) = P(f((d(x)) = y | f(x) = y)$, where $y$ is the ground truth. We consider two kinds of transformations: 1) Synthetic distortions, such as adjusting the color saturation of an image, adding noise, etc., and 2) natural transformations that exist across consecutive frames of videos.

Unlike some prior definitions of model robustness, our robustness definition is agnostic to the kind of transformation function. In \cite{hendrycks2018benchmarking}, robustness according to the relative mCE is defined to be proportional to $P(f(x) = y) - P(f(d(x)) = y)$. This formulation implicitly assumes that transformed images come from different distribution than the clean images, and are harder to classify. This robustness metric is trivial when considering natural robustness, because neighboring frames come from the same distribution and are equally difficult to classify.

\section{Experimental Study Setup}
\label{study_setup}
\subsection{Dataset}
We leverage a large scale dataset of videos, YouTube-BoundingBoxes(YT-BB) (\cite{real2017youtube}), to evaluate natural robustness. YT-BB contains 380K video segments of 15-20s across 210K unique videos and 24 classes. Each segment is selected to feature objects in natural settings without editing or post-processing. See Figure \ref{fig:example_data} for examples of both natural transformations and synthetic distortions. 

\subsection{Task}
We create a classification task from YT-BB to train our image classification model. This ensures domain matching between images used during training and the video frames used during robustness evaluation. Prior work used handpicked YouTube videos to evaluate image models trained on ImageNet (\cite{polar_bear}). However, domain mismatch between training and evaluation can result in a significant drop in test accuracy. In our evaluation, we found that evaluating off-the-shelf ImageNet models on the YT-BB classification tasks, without finetuning, results in an average of 27\% drop in accuracy across 12 model architectures.

To create this classification task, we split 210K videos into training, validation, and test. We divide the videos into contiguous shots of a single object, and randomly sample a subset of 56,209 frames from unique shots, which we term "anchor frames". These anchor frames make up the supervised classification task across 23 imbalanced classes. We then extract 5 consecutive frames at 15Hz before and after the anchor frames to evaluate natural robustness. The top row of figure \ref{fig:example_data} illustrates an example anchor frame and its neighbors, which we term "natural transformations". Neighboring frames are only used to evaluate natural robustness. Only anchor frames are used during training. For reproducibility, we release the video robustness dataset with exact anchor frames used in train, validation, and test in the ArXiv directory.

\subsection{Models}
We examine the robustness of 12 model architectures: VGG16, VGG19 (\cite{vgg-simonyan2014very}), ResNet-V1-50, ResNet-V1-101, ResNet-V1-152 (\cite{resnet-he2016deep}), MobileNet-V1 (\cite{howard2017mobilenets}), NASNet-Mobile (\cite{nasnet-zoph2018learning}), Inception-V1 (\cite{inception1-szegedy2015going}), Inception-V2, Inception-V3 (\cite{inception23-szegedy2016rethinking}), Inception-V4 (\cite{inception4-szegedy2017inception}), and Inception-ResNet-V2 (\cite{inception-resnet-szegedy2017inception}). For each model architecture, we train models on the YT-BB task from scratch and using transfer learning from pre-trained ImageNet checkpoints. We tune hyper-parameters using the same search space for every model independently. This gives us a total of 24 baseline models with a wide range of accuracy on this task. 

The preprocessing stage during training is particularly critical in determining the robustness of models against synthetic distortions. \cite{geirhos2018generalisation} demonstrated that training on synthetic distortions will disproportionately improve robustness to the distortion used during training. To ensure that our evaluation across models are fair and comparable, we use the same preprocessing step during all model training, using only random crop and resizing, skipping the color distortions that is typically used for training Inception models.

\subsection{Natural Transformations and Synthetic Distortions} 
For each anchor frame in our classification task, we sample 10 neighboring frames, 5 frames before and after the anchor frames, at 15Hz. We call these neighboring frames the natural transformations of the anchor image. Similar to synthetic distortions, we can adjust the strength of the transformation defined by the temporal difference between the anchor and neighboring frame. 

We also evaluated model robustness to synthetic distortions for comparison against natural robustness. We implemented 10 varieties of synthetic distortions at 5 different levels of severity. We selected the following synthetic distortions from prior work: gaussian noise, gaussian blur, pixelate, shot noise, JPEG quality, hue, contrast, saturation, brightness, and small translations (\cite{hendrycks2018benchmarking, polar_bear, geirhos2018generalisation}). 

\cite{polar_bear} showed that the implementation of the translation distortion can significantly affect the results. In our implementation, we shift the evaluation crop of the image to create a set of translations, without introducing empty space or the need for in-painting.

\section{Results}

We examine the result of evaluating each of the 24 fully trained models, of 12 model architectures, on 11 transformations, each with 5 varying strengths. For the 12 models trained from scratch, the Top-1 accuracy on the 23-class YTBB classification task ranges from 72.5\% to 80.4\%. The same architectures fine-tuned from ImageNet obtained accuracy ranging from 83.90\% and 88.7\%. 

\subsection{Frames temporally closer in time have more similar model outputs}

\begin{figure}[h]
\begin{center}
\includegraphics[width=\columnwidth]{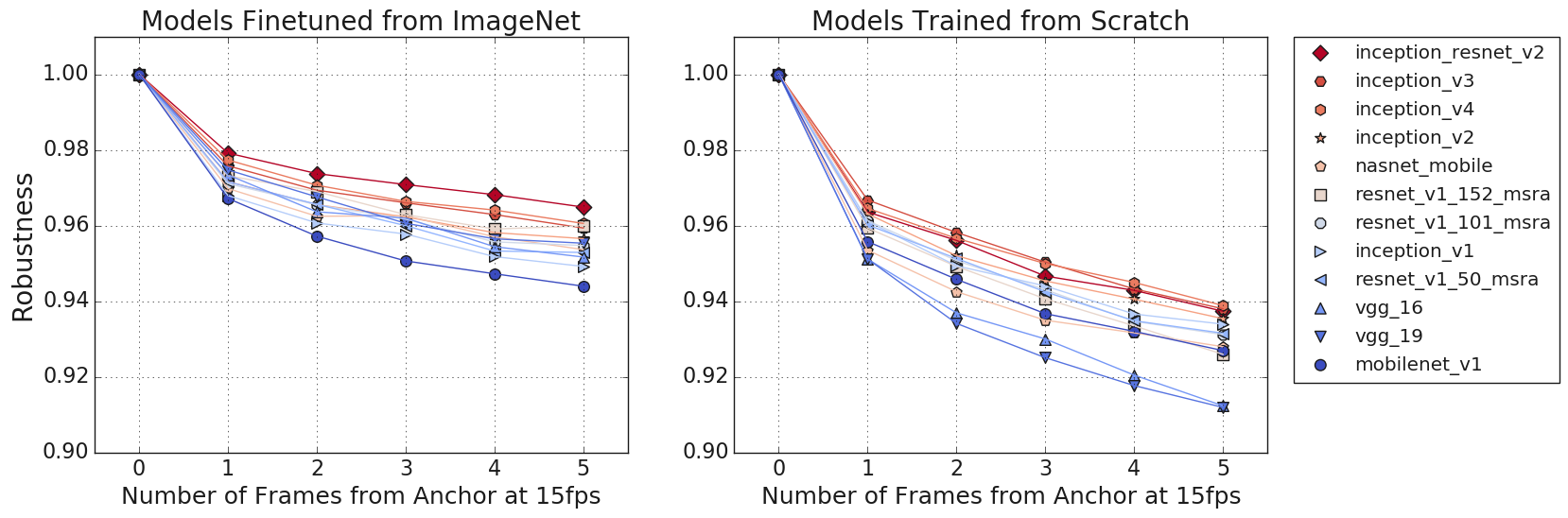}
\end{center}
\caption{
{\bf Video frames that are closer in time (more visually similar) are more likely to result in similar model predictions.} We plot natural robustness as a function of the time elapsed between two frames, conditioned on the first frame being classified correctly. Model colors are determined by their test accuracy, ranging from 88.7\% (Inception-ResNet-V2) to 83.9\% (MobileNet-V1), when finetuned from ImageNet. Left: models fine-tuned on YT-BB task from ImageNet. Right: models trained on YT-BB task from scratch. 
}
\label{fig:robustness_over_time}
\end{figure}

We expect our models to exhibit the property that more visually similar images have more similar outputs. During our evaluation, we found that this is indeed true. As seen in Figure \ref{fig:robustness_over_time}, frame pairs temporally closer together are more likely to be both classified correctly. 

We also evaluated the robustness of YT-BB models on synthetic distortions. The result can be found in \ref{fig:synth_robustness_over_time}. Our result reproduces that of \cite{geirhos2018generalisation}, showing that models are not robust to synthetic distortions and that robustness drops significantly as the severity of distortions increase.

\subsection{More accurate models are more robust}

\begin{figure}[h]
\begin{center}
    \includegraphics[width=\columnwidth]{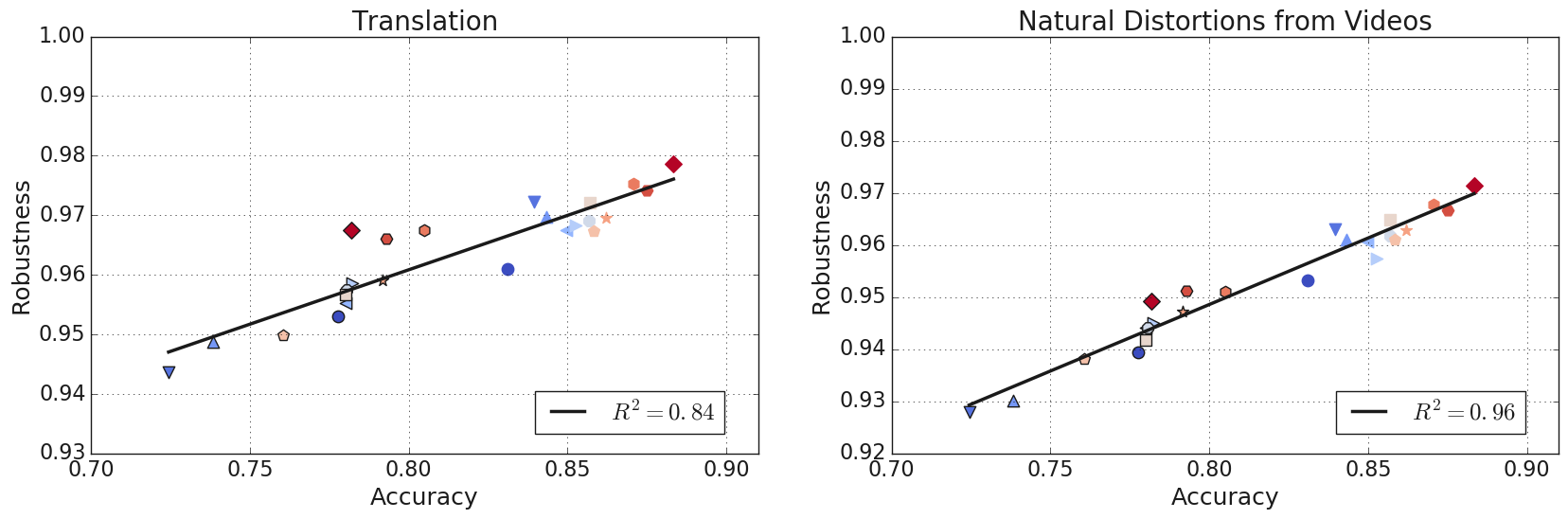}
\end{center}
\caption{
{\bf More accurate models are generally more robust.} We plot the average robustness to two in-distribution transformations: translation (left) and natural transformation (right) against model accuracy. Robustness measure is averaged across 5 varying strengths of distortions. See Figure \ref{fig:robustness_over_time} for a common color legend. Models fine-tuned from ImageNet are borderless. 
}
\label{fig:natural_robustness}
\end{figure}
\begin{figure}[h]
\begin{center}
    \includegraphics[width=4.5in]{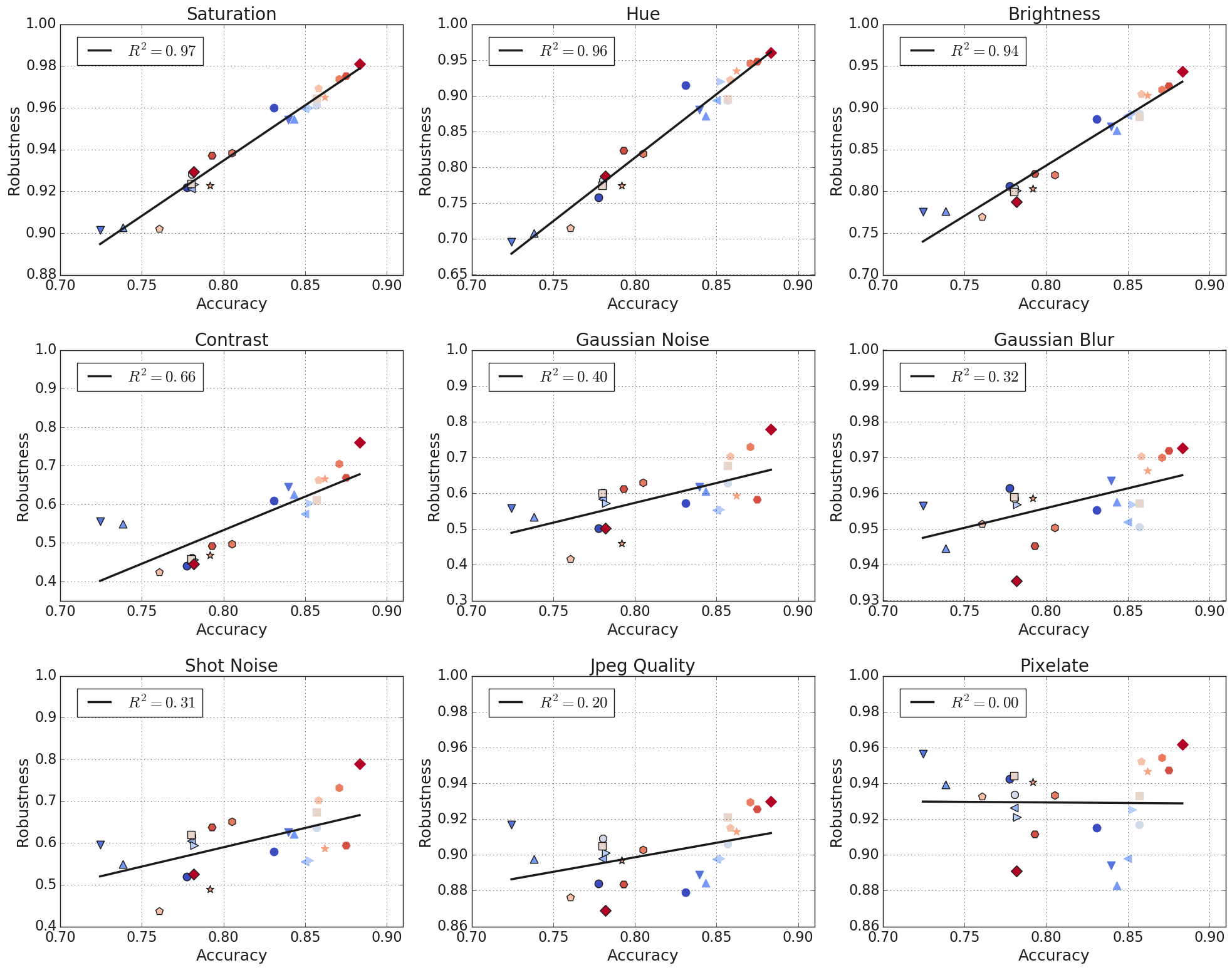}
\end{center}
\caption{
{\bf Model accuracy on the test set is positively correlated with model robustness to synthetic color distortions  but not correlated to robustness to other noisy synthetic distortions}. Plots are ordered based on $R^2$. Note that all models are with only random crop and resizing during preprocessing. See Figure \ref{fig:robustness_over_time} for a common legend. Models fine-tuned from ImageNet are borderless.
}
\label{fig:synthetic_robustness}
\end{figure}

We summarize each model's robustness to a given transformation as the average across varying strengths. In Figure \ref{fig:natural_robustness}, we see a strong correlation between model accuracy and robustness to both translation and natural transformations found in videos. The result on translation robustness contradicts the conclusion of \cite{polar_bear}, which states modern networks are less translation invariant. \cite{polar_bear} measure translation invariance by training their models on full-sized images, then evaluating their models by embedding the image within a larger image. This introduces a large change between training and evaluation, and Figure 8 in \cite{polar_bear} indicates the closer embedded size is to the original image, the more robust models are. Our work is focused on evaluating robustness to small transformations that are close to or even in the same distribution as those seen in training.

Figure \ref{fig:synthetic_robustness} shows that more accurate models are also more robust to a number of synthetic distortions such as saturation, hue, and brightness. This conclusion is different from that of \cite{hendrycks2018benchmarking}, which claimed that model robustness is largely uncorrelated with model accuracy, according to their relative mCE metric. However, according to their unadjusted mCE robustness metric, more accurate classifiers are more accurate on the corrupted test set. As described in Sec. \ref{sec:metric}, the relative mCE metric is proportional to the drop in total accuracy from clean images to corrupted images. Our metric performs a more fine-grained measure of robustness, by computing accuracy on the corrupted image conditional on the clean image being classified correctly, rather than simply using the total accuracies.

\subsection{Synthetic distortions as proxies of natural transformations}

\begin{figure}[h]
\begin{center}
\includegraphics[width=3in]{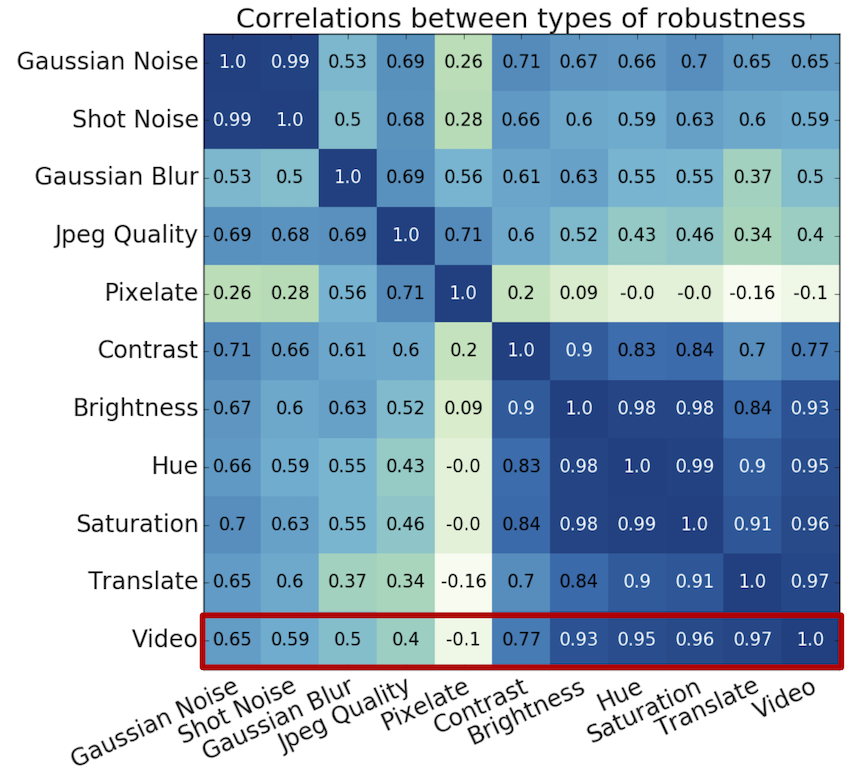}
\end{center}
\caption{
{\bf Image translation, and synthetic adjustments to saturation and hue are good proxies for evaluating natural robustness.} We evaluate the correlation between pairs of robustness types to identify easy-to-implement proxies for evaluating natural robustness. Each pair of correlations is computed over 24 fully trained and tuned models across 12 model architectures.
}
\label{fig:correlations}
\end{figure}

In many settings, it's difficult to acquire domain-matching videos to evaluate natural robustness. We evaluate model robustness across 10 different synthetic distortions, and compute the correlation between different types of robustness. Shown in Figure \ref{fig:correlations}, we find that robustness to image translation and color distortions like saturation and hue are highly correlated with natural robustness, which indicates that these are good proxies for natural robustness in image models.

\subsection{Distances between neighboring frames are much larger than distances between adversarial examples}

\begin{figure}[h]
\begin{center}
    \includegraphics[width=2.5in]{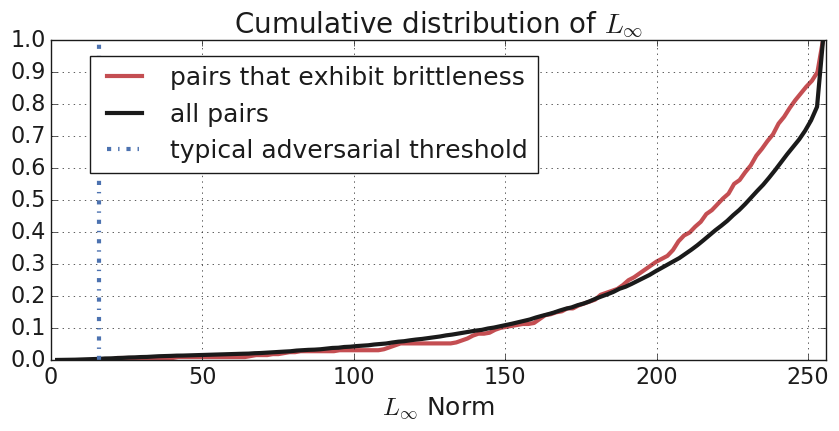}
\end{center}
\caption{
{\bf Distances between neighboring frames are larger than distances between typical adversarial examples. The same holds for neighboring frames that exhibit brittleness.} We plot the cumulative distribution (CDF) of $L_\infty$ norm for all frame pairs 66ms apart (black), and the CDF for pairs that exhibit brittleness. Only $1.6\%$ of normal frame pairs and less than $0.01\%$ of frame pairs that exhibit brittleness are found in the typical adversarial examples neighborhood. 
}
\label{fig:adversarial}
\end{figure}

We next investigated the relationship between brittleness to natural transformations and adversarial examples. Adversarial examples are commonly defined as images within an $\epsilon-$ball of the clean image by $L_\infty$ norm that results in a mis-classification. 

In Figure \ref{fig:adversarial}, we analyze the distribution of 10,000 video frame pairs, all 66ms apart. The distribution of distances between consecutive frames have a mean of 213 and standard deviation of 49.1, which is much larger than what is typically considered for adversarial examples (less than an $\epsilon=16$ in $L_\infty$ space (\cite{kannan2018adversarial}). We then study frame pairs that exhibit brittleness, where the anchor frame is classified correctly but its neighbor is mistaken. Less than 0.01\% of frame pairs that exhibit brittleness fall within the definition of adversarial examples ($\epsilon=16$ ball in $L_\infty$ space).

\subsection{No single training technique systematically improves robustness}
\label{sec:enhancement}

\begin{figure}[h]
\begin{center}
\includegraphics[width=\columnwidth]{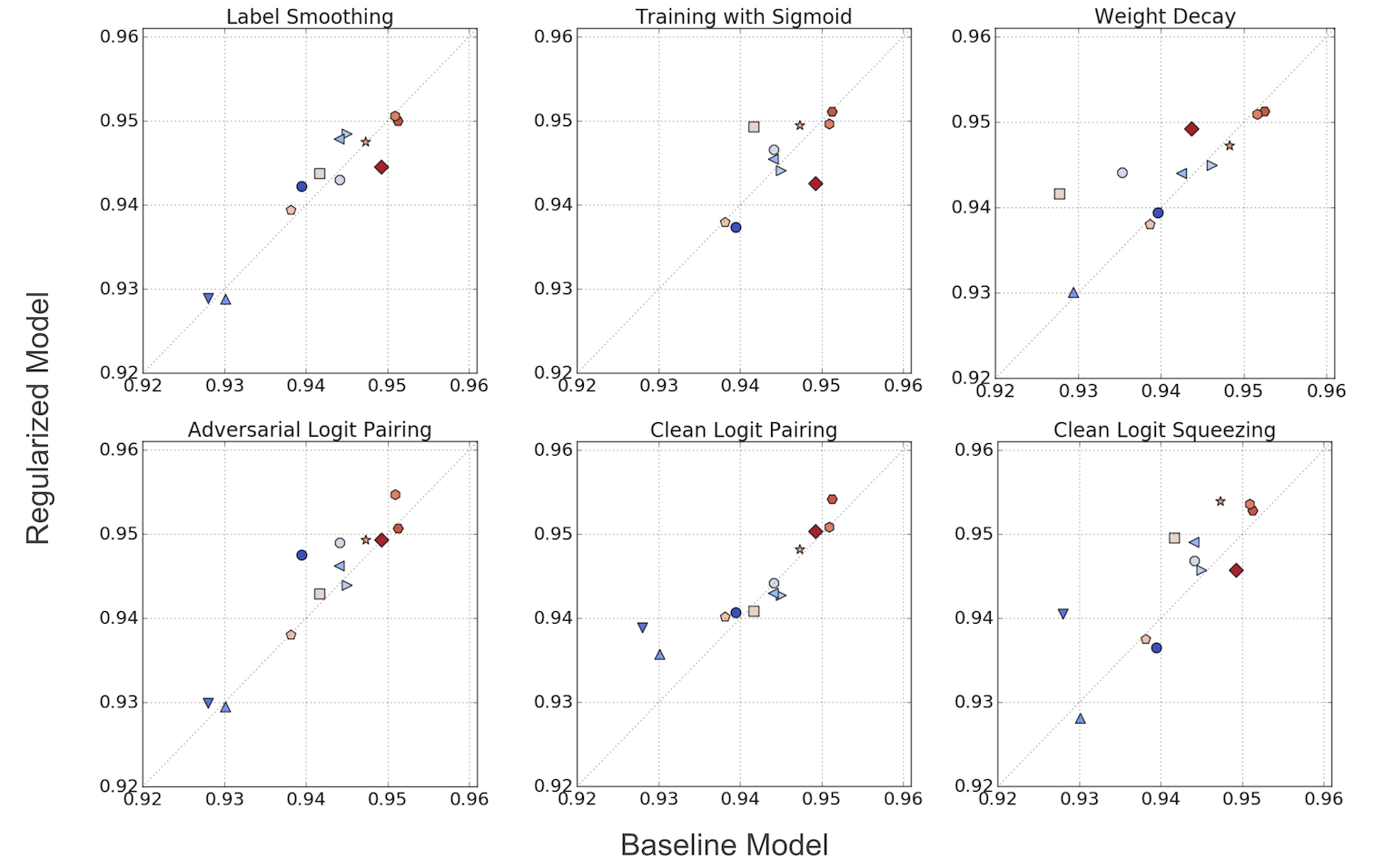}
\end{center}
\caption{
{\bf No single adversarial training or regularization technique systematically improves robustness across model architectures}. For each regularization technique, we plot the robustness after regularizing against robustness of the baseline model. We plot the line of equality to show whether regularizing improves robustness. See Figure \ref{fig:natural_robustness} for common legend. 
}
\label{fig:rvr}
\end{figure}

Finally, we explore regularization and adversarial training techniques to improve the natural robustness of image classifiers. We explored adversarial training through adversarial logit pairing (\cite{kannan2018adversarial}); regularization techniques like weight decay (\cite{wd-hanson1989comparing}), label smoothing (\cite{inception4-szegedy2017inception}), clean logit squeezing, clean logit pairing (\cite{kannan2018adversarial}); and multi-class prediction with the sigmoid activation function on the logits (\cite{goodfellow2016deep}). We explore 25 hyperparameter settings for each training technique. All models trained are within 1.2\% of their original accuracy.

We find that no single technique we explored improves natural robustness for all model architectures (Figure \ref{fig:rvr}). However, certain model architectures become more robust with specific training techniques. Figure \ref{fig:rvr} shows that adversarial logit pairing improves MobileNet-V1 robustness by 0.8\%, weight decay improves Resnet-V1-152 robustness by 1.2\%, and clean logit pairing and clean logit squeezing improves VGG19 robustness by 1\% and 1.3\% respectively.

\section{Conclusion}
In this work, we present the first study on natural robustness in CNNs by leveraging the large YT-BB video dataset (\cite{real2017youtube}). 

Our results show that more accurate models are also more robust to natural transformations. This implies that researchers should continue designing new architectures and training techniques that improve model accuracy. Our analysis also highlights that the correlation between accuracy and robustness depends highly on the type of distortions applied. This refines conclusions from prior works that show accuracy is uncorrelated with robustness on average (\cite{hendrycks2018benchmarking}). 

When video frames are not available to evaluate natural robustness directly, we identify synthetic color distortions as good proxies for natural transformations. This result also helps explain and support prior findings that color-based transformations are good data augmentation policies for natural images (\cite{cubuk2018autoaugment}). 

In exploring the relationship between brittleness found in videos and adversarial examples, we find that brittle examples in videos rarely fall within the typical definition of adversarial examples. This suggests that adversarial robustness does not directly measure robustness to natural transformations. Despite the misalignment in evaluation, we do find early signs that training techniques to improve adversarial robustness can improve robustness for some model architectures. However, no single training technique systematically improves natural robustness across model architectures, providing an interesting direction for future work.


\bibliography{iclr2019_conference}
\bibliographystyle{iclr2019_conference}

\appendix

\section{Appendix}
\renewcommand\thefigure{\thesection.\arabic{figure}}    
\setcounter{figure}{0}    

\begin{figure}[h]
    \begin{center}
        \includegraphics[width=\columnwidth]{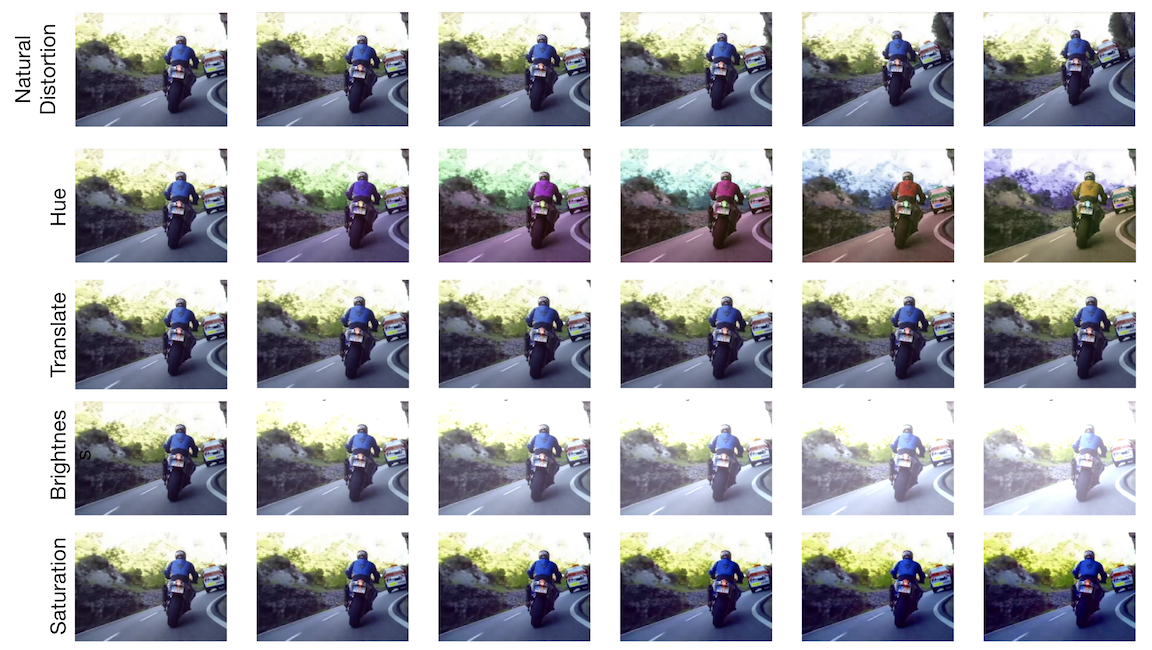}
    \end{center}
\caption{
    {\bf Examples of natural transformations found in videos and synthetic distortions.} From the publicly available YT-BB video dataset, we sample anchor frames from video segments that are at least 666ms long. Top row: Neighboring frames are used to evaluate models' natural robustness. Remaining rows: Examples of of synthetic distortions. From left to right, we show distortions with increasing strength. 
    }
\label{fig:example_data}
\end{figure}

\begin{figure}[h]
\begin{center}
    \includegraphics[width=2.5in]{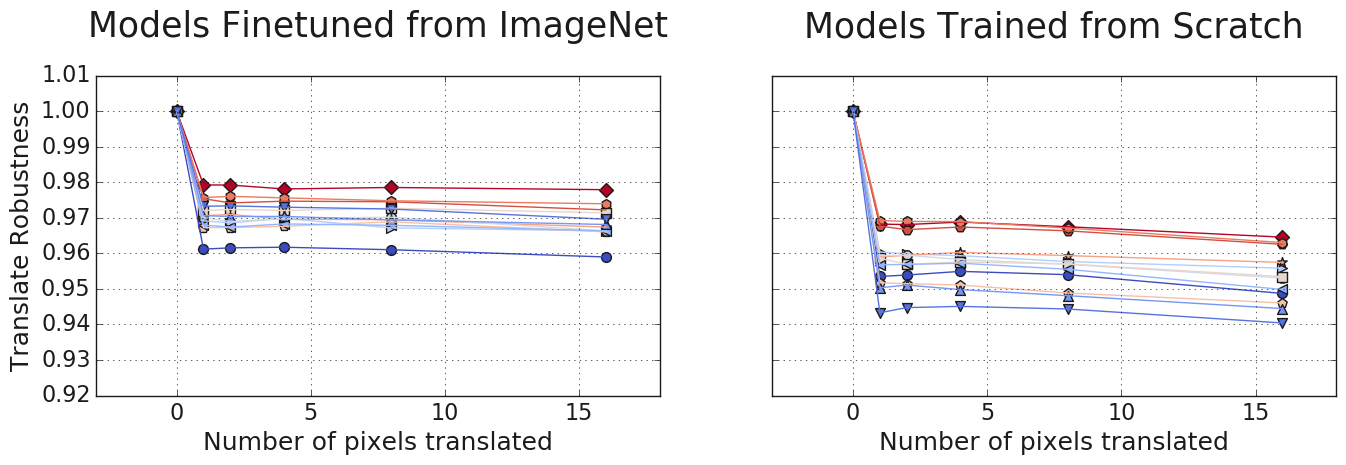} \includegraphics[width=2.5in]{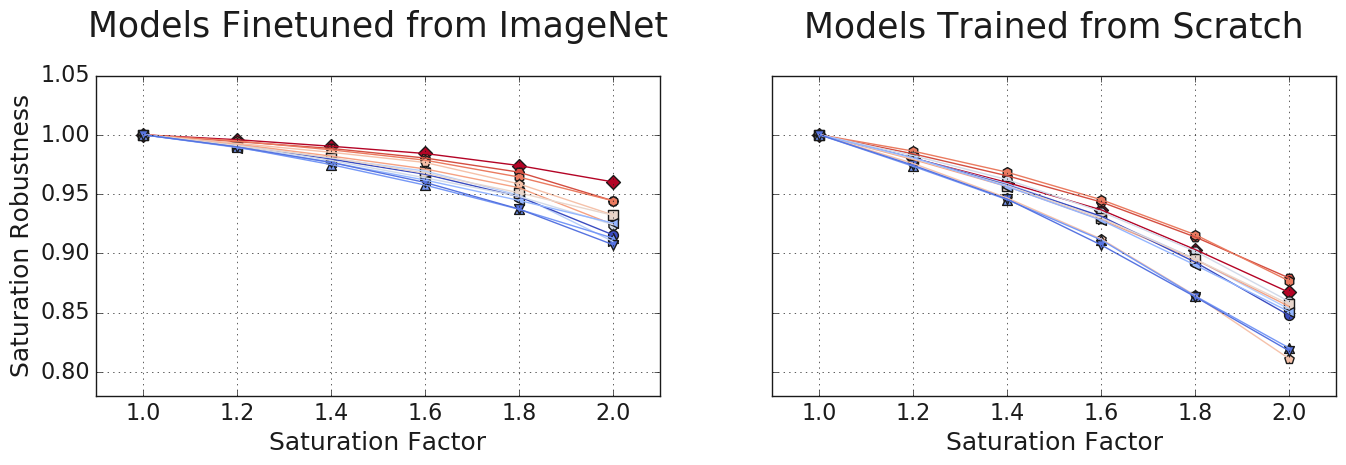}
    \includegraphics[width=2.5in]{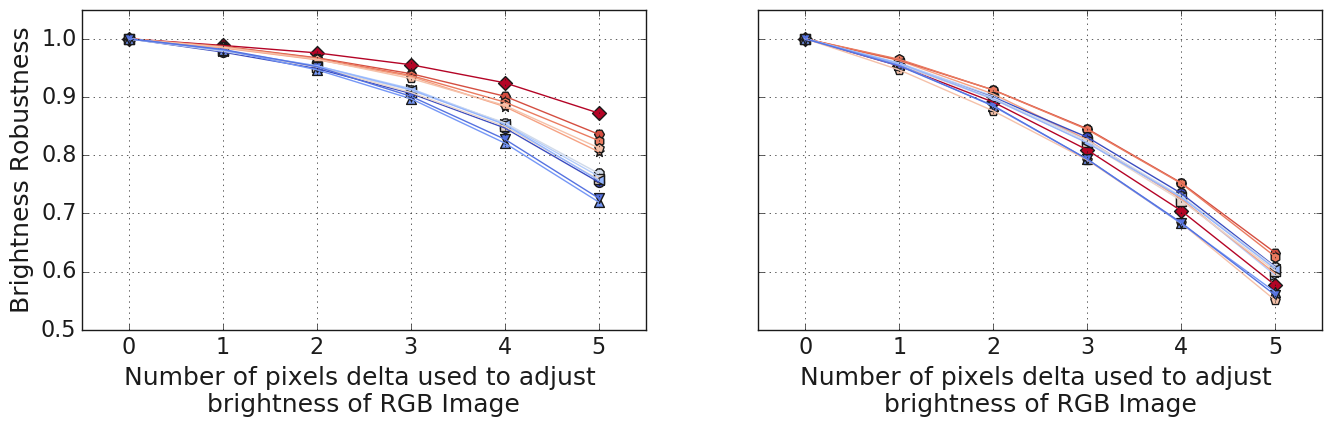} \includegraphics[width=2.5in]{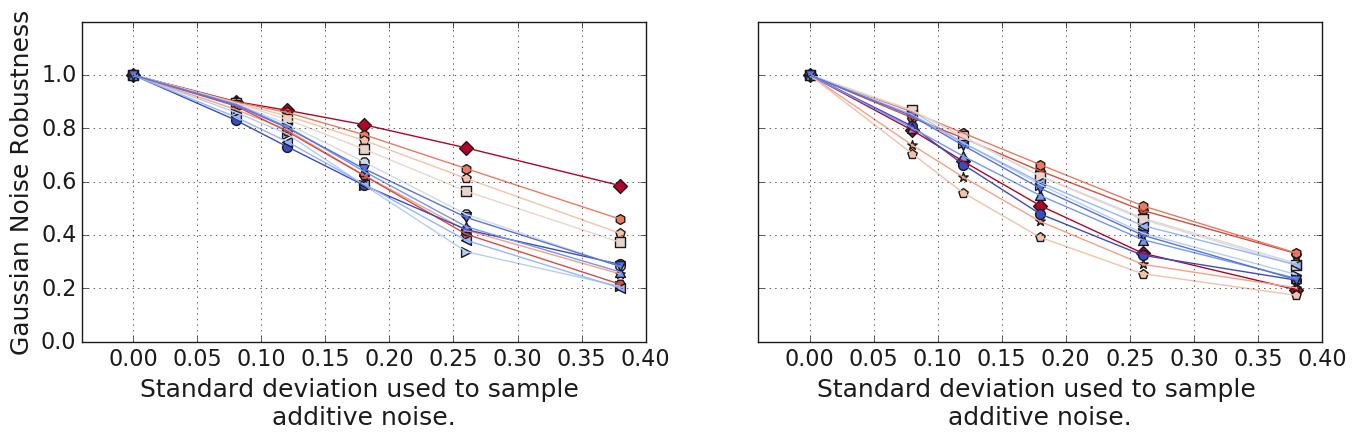}
    \includegraphics[width=2.5in]{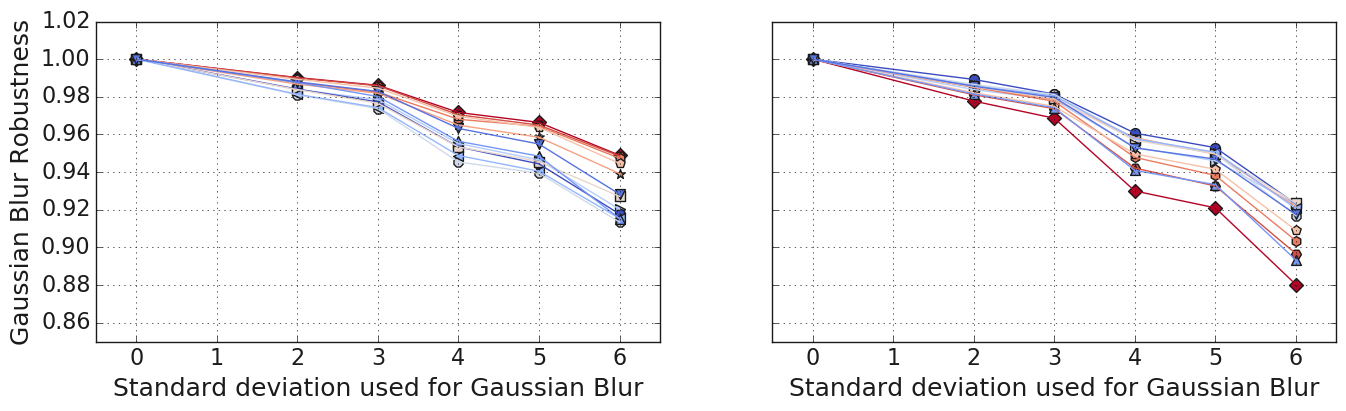}    \includegraphics[width=2.5in]{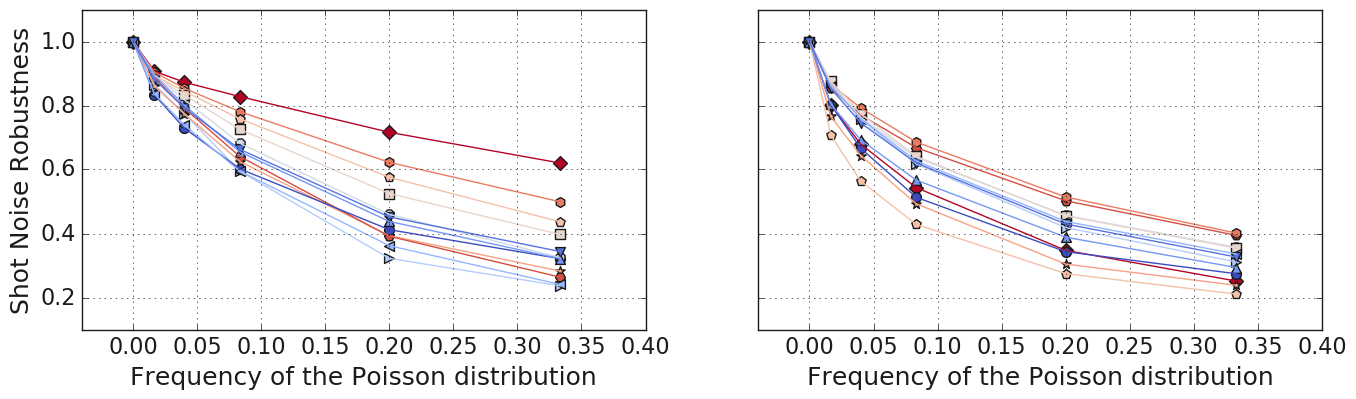}
    \includegraphics[width=2.5in]{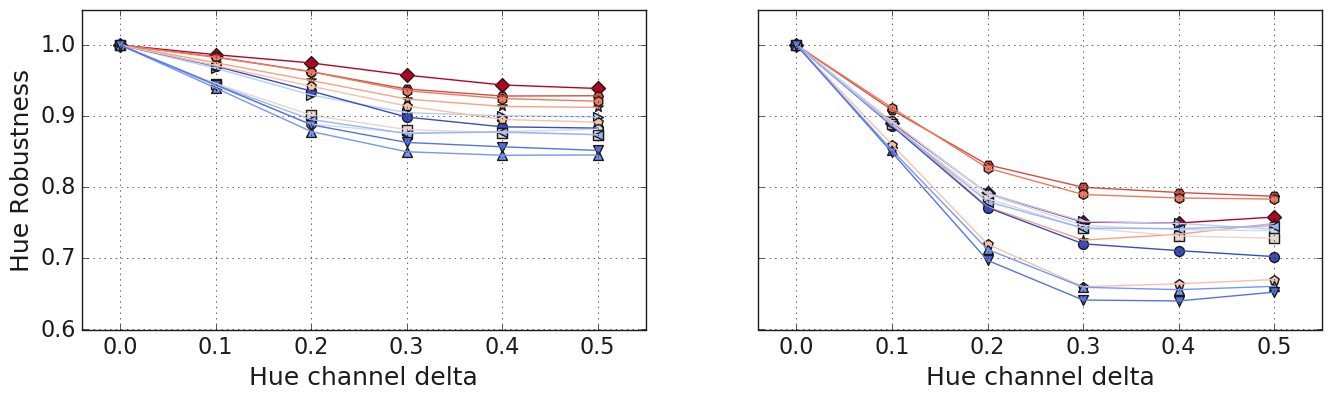}    \includegraphics[width=2.5in]{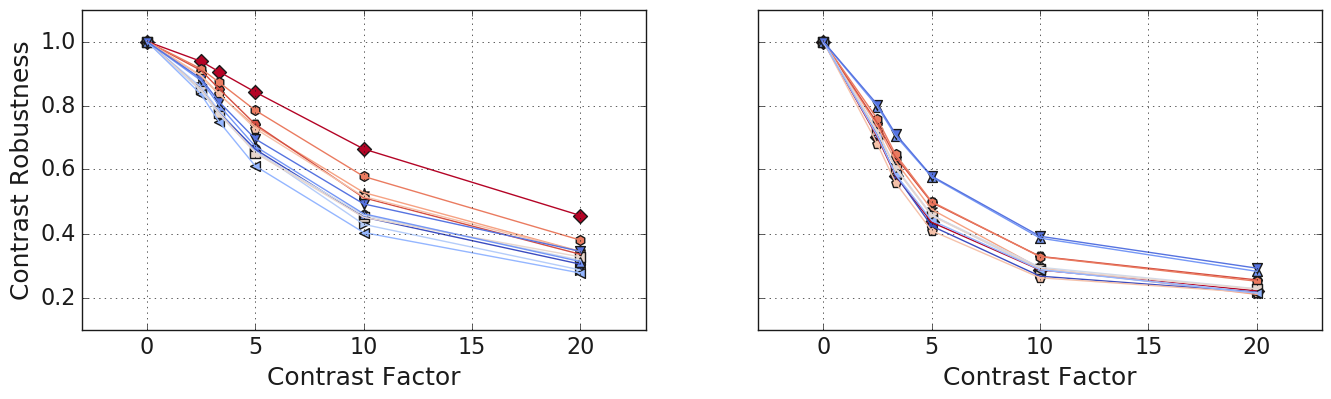}
    \includegraphics[width=2.5in]{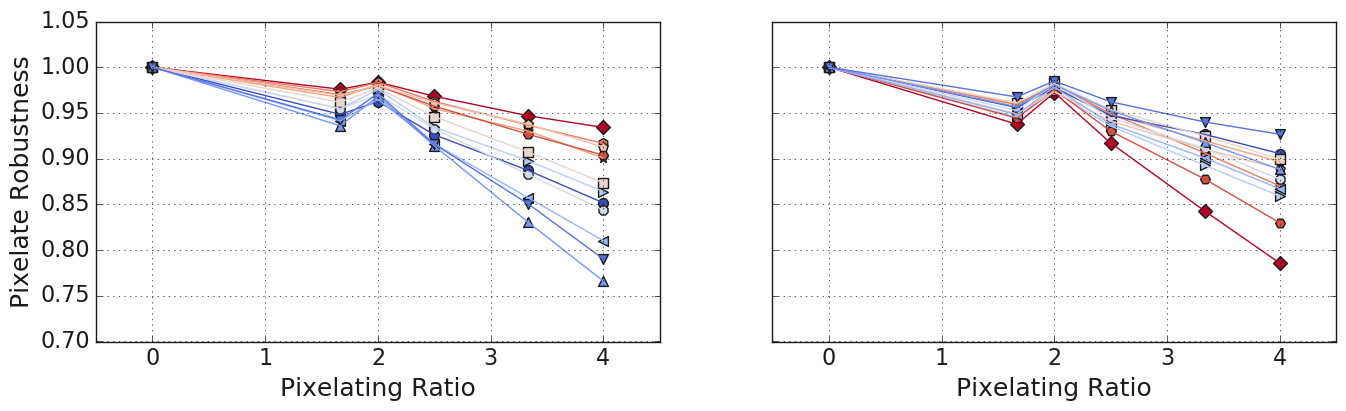}    \includegraphics[width=2.5in]{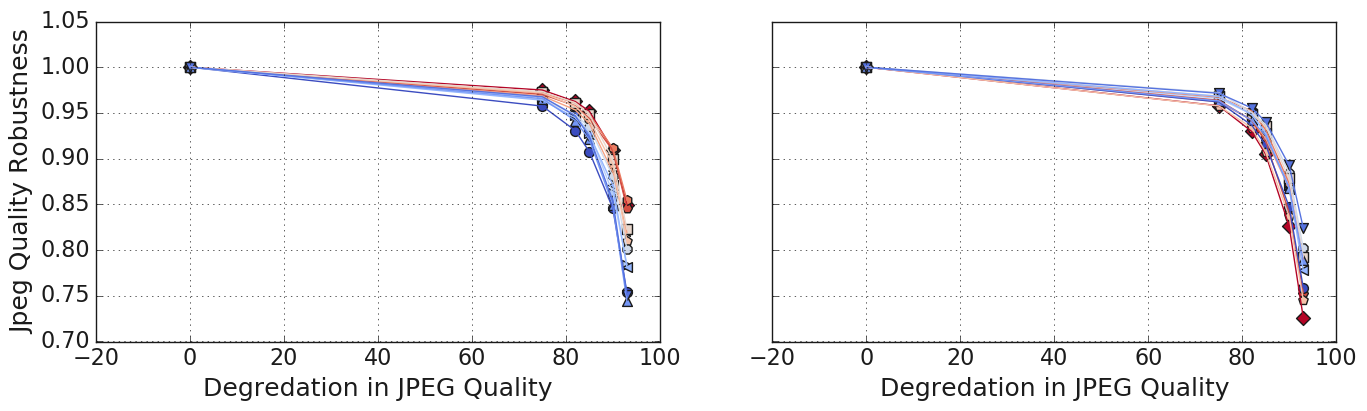}
    
\end{center}
\caption{
{\bf Increasing the strength of synthetic distortions reduces robustness}, as previously shown in \cite{geirhos2018generalisation}. Similar to Figure \ref{fig:robustness_over_time}, we plot the robustness of models with respect to varying strength of synthetic distortions.
}
\label{fig:synth_robustness_over_time}
\end{figure}

In Figure \ref{fig:synth_robustness_over_time}, we illustrate the robustness across degrees of synthetic distortions, replicating previous results from \cite{geirhos2018generalisation}. Robustness to small translations stands out from other types of robustness in that models are equally robust to translations of a single pixel or 16 pixels. We hypothesize this is due to the use of random crops during training. Additionally, we notice that models are more robust to a translation of exactly 4 and 8 pixels as opposed to 1 or 16 pixels. This phenomenon is likely due to the convolution architecture, as alluded to by  \cite{polar_bear}.

\end{document}